\documentclass{article}

\usepackage{PRIMEarxiv}
\usepackage[utf8]{inputenc} 
\usepackage[T1]{fontenc}    
\usepackage{hyperref}       
\usepackage{url}            
\usepackage{booktabs}       
\usepackage{amsfonts}       
\usepackage{nicefrac}       
\usepackage{microtype}      
\usepackage{lipsum}
\usepackage{fancyhdr}       
\usepackage{graphicx}       
\usepackage[title]{appendix}
\usepackage[square,numbers]{natbib}
\usepackage{float}
\usepackage{amsthm}
\usepackage{caption}
\usepackage[framemethod=tikz]{mdframed}
\usepackage{etoolbox} 
\usepackage{enumitem}
\usepackage{orcidlink}
\usepackage{array}
\usepackage{multirow}
\usepackage[autostyle]{csquotes}  
\graphicspath{{media/}} 

\newmdtheoremenv[
  hidealllines=true,
  leftline=true,
  innerleftmargin=10pt,
  innerrightmargin=10pt,
  skipabove=10pt,
  skipbelow=10pt,
]{prompt}{Prompt}

\newmdtheoremenv[
  hidealllines=true,
  leftline=true,
  innerleftmargin=10pt,
  innerrightmargin=10pt,
  skipabove=10pt,
  skipbelow=10pt,
]{output1}{Output}

\makeatletter

\title{Imagining from Images with an AI Storytelling Tool}

\author{
  Edirlei Soares de Lima \orcidlink{0000-0002-2617-3394}\\
  Academy for AI, Games and Media \\
  Breda University of Applied Sciences \\
  Breda, The Netherlands\\
  \texttt{soaresdelima.e@buas.nl} \\  
  \And
  Marco A. Casanova \orcidlink{0000-0003-0765-9636}\\
  Department of Informatics \\
  PUC-Rio \\
  Rio de Janeiro, Brazil\\
  \texttt{casanova@inf.puc-rio.br} \\  
  \And
  Antonio L. Furtado \orcidlink{0000-0003-3710-624X}\\
  Department of Informatics \\
  PUC-Rio \\
  Rio de Janeiro, Brazil\\
  \texttt{furtado@inf.puc-rio.br} \\
}

\begin{document}
\maketitle

\begin{abstract}
A method for generating narratives by analyzing single images or image sequences is presented, inspired by the time immemorial tradition of Narrative Art. The proposed method explores the multimodal capabilities of GPT-4o to interpret visual content and create engaging stories, which are illustrated by a Stable Diffusion XL model. The method is supported by a fully implemented tool, called ImageTeller, which accepts images from diverse sources as input. Users can guide the narrative's development according to the conventions of fundamental genres — such as Comedy, Romance, Tragedy, Satire or Mystery —, opt to generate data-driven stories, or to leave the prototype free to decide how to handle the narrative structure. User interaction is provided along the generation process, allowing the user to request alternative chapters or illustrations, and even reject and restart the story generation based on the same input. Additionally, users can attach captions to the input images, influencing the system's interpretation of the visual content. Examples of generated stories are provided, along with details on how to access the prototype.
\end{abstract}


\section{Introduction}

Images are, first of all, \textit{descriptive}. A landscape describes a place, a portrait describes the traits of a person. But they can also be \textit{narrative}, if they display an event occurring in a given place and enacted by one or more persons. When a sequence of two or more images is presented, such as step-by-step successive scenes constituting an event, the narrative aspect becomes even more evident.

In this paper, we propose to harness, in a prototype named \textit{ImageTeller}, the narrative power of images and image sequences to compose stories inspired by the portrayed scenes, taking advantage of recent advancements in multimodal models that integrate text, vision, and audio capabilities, such as GPT-4o.\footnote{\href{https://openai.com/index/hello-gpt-4o/}{https://openai.com/index/hello-gpt-4o/}}

\begin{figure}[htb]
	\centering
	\includegraphics[width=0.25\linewidth]{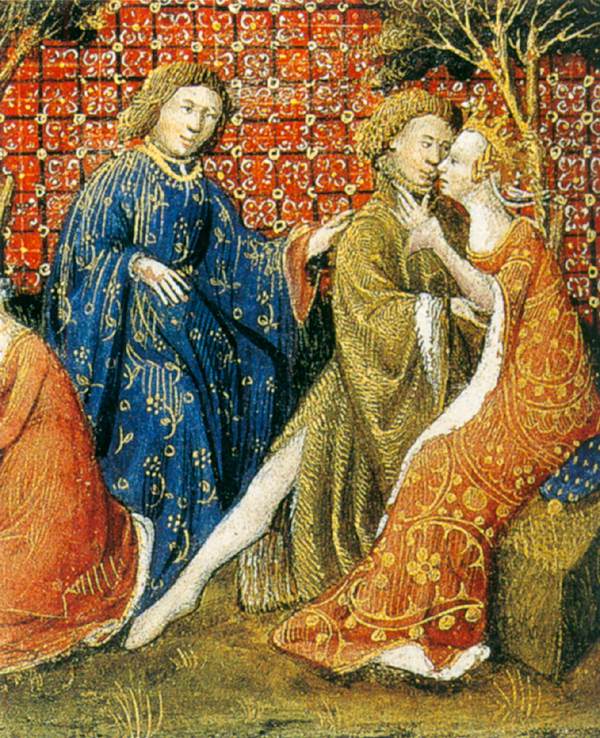}
	\caption{First tryst of Guinevere and Lancelot, arranged and observed by Galehaut \cite{lancelot1470}.}
	\label{fig1}
\end{figure}

The popular adage that \enquote{a picture is worth a thousand words} is not only true in the sense that images clarify what may be difficult to understand in a textual explanation,\footnote{Of which Figure \ref{fig2} in Section \ref{sec3} provides a poignant example, as an indispensable help to the reader of this paper when trying to decipher how the proposed prototype works.} but also in view of the rich variety of widely different stories that come to the imagination when contemplating the images. Figure \ref{fig1} shows a medieval illustration where Guinevere and Lancelot appear kissing each other. The person dressed in blue is the knight Galehaut, who arranged this first tryst, and his unrealistic presence contributes to narrate an entire love episode in a single compact scene. Among the early examples in Section \ref{sec4}, we explore how this love scene in turn generated the tragic story of Francesca da Rimini and Paolo Malatesta in Dante Alighieri's \textit{Divine Comedy}.

The images accepted as input by our ImageTeller prototype can originate from various media, such as books, newspapers or article illustrations, comic strip cartoons, paintings, drawings, sculptures, photographs, scenes captured from movies and videos, tables and diagrams in business reports, etc., provided they are stored in an image file format.

Input image sequences of two or more frames may come from the same source, implicitly representing a real or fictional event, but may also come from different sources, which adds a layer of complexity for the prototype in its aim to devise a novel coherent narrative. Successful stories generated by the system should ideally enable the user to feel that they are indeed consistent with what the input sequences suggest.

The proposed narrative generation process allows the user to either structure the plot according to the conventions of one of the five fundamental genres that we have previously specified, namely \textit{comedy}, \textit{romance} (in the original sense of epic plots, flourishing since the twelfth century), \textit{tragedy}, \textit{satire}, \textit{mystery} \cite{delima2024multigenre} or to select a data storytelling genre. Alternatively, the user may leave the prototype free to determine how to handle each chapter and the overall plot direction. The system also supports user interaction throughout the generation process, allowing the user to call for alternative chapters or illustrations and even restart the entire process using the same input. A subtle feature for user interaction is the optional ability to attach \textit{captions} to the input frames, influencing how each image should be interpreted.

The concept of generating stories from visual input has been explored in a few previous related works. Early research in this area, such as the work by Farhadi et al. \cite{farhadi2010}, focused on generating descriptive sentences from images by comparing estimations of meaning derived from both visual and textual data. Subsequent studies explored the potential of AI models to create more elaborate stories based on images or image sequences. For instance, Huang et al. \cite{huang2016} proposed a system that uses machine learning to generate descriptive captions for images and then constructs short stories from these captions. In a similar work, Smilevski et al. \cite{smilevski2018} introduced a Sequence-to-Sequence model with separate encoders for visual and narrative components, aiming to generate longer, more human-like stories that go beyond basic image descriptions. Addressing the need for discourse coherence in visual storytelling, Cardona-Rivera and Li \cite{cardona2021} introduced PlotShot, a system that generates stories around user-supplied photographs by considering discourse constraints during fabula generation. While these approaches have shown promise in automated storytelling, they often focus on generating short, descriptive narratives or captions rather than longer, structurally complex stories. ImageTeller builds upon these previous efforts by exploring the multimodal capabilities of GPT-4o and its vision counterpart to generate richer and more engaging stories from images.

The paper is structured as follows. Section \ref{sec2} offers a brief review of concepts from the topic of \textit{Narrative Art}, which we found necessary to establish a foundation for this work. Section \ref{sec3} provides full technical details about the architecture, functionalities, and usage of ImageTeller, including links to access the prototype. Section \ref{sec4} reports the experiments conducted with ImageTeller. Section \ref{sec5} presents concluding remarks.

\section{On the Narrative Art Tradition}
\label{sec2}

In literary works, such as novels or poems, the text is the dominant component, whereas images, when inserted as \textit{illustrations}, play the double secondary role of inducing readers to mentally visualize the textual content, while serving to embellish the printed pages. In both roles, some famous illustrators, like Gustave Doré, have added an enduring seductive enchantment to classic masterpieces, notably Dante's \textit{Divine Comedy} and Perrault's fairy tales.

However, images are known to have long preceded writing, given that \textit{Narrative Art} \footnote{\href{https://en.wikipedia.org/wiki/Narrative\_art}{https://en.wikipedia.org/wiki/Narrative\_art}} has a tradition whose beginning is continuously being traced to earlier times. In a recent \textit{Nature} article \cite{oktaviana2024}, we learn of a \enquote{newly described cave art scene at Leang Karampuang}, which is a limestone cave situated in the island of Sulawesi (a.k.a. Celebes), in Indonesia: \enquote{Painted at least 51,200 years ago, this narrative composition, which depicts human-like figures interacting with a pig, is now the earliest known surviving example of representational art, and visual storytelling, in the world.}

The use of Narrative Art was widespread in ancient civilizations, including those of Mesopotamia, Egypt, Greece, and Rome \cite{stokstad2017}. As evidenced by extant monuments and fragmentary records, these cultures employed visual imagery to tell stories about their gods, heroes, and historical events, often incorporating inscriptions to further enhance the narrative.

Medieval manuscripts later gave a special impulse to the tradition, interspersed as they often were with narrative \textit{illuminations} that occasionally help to recover information on legendary tales, or even specific legendary objects of central significance, such as the Grail. A creation of the French writer Chrétien de Troyes, the Grail's nature was never fully revealed since the manuscript was not finished, conceivably due to the author's premature death. No less than five \enquote{Continuations} were written by other authors in an attempt to solve the mystery, being available today, together with the initial Chrétien's romance, in voluminous book format \cite{chretien2015}. But what matters to the present discussion is that comparing the different shapes of the Grail, taken in a sizable number of illuminations, has suggested an alternative strategy to investigate the enigmatic nature of the object \cite{meuwese2008}.

Still in Roman times, Virgil's \textit{Iliad} similarly dedicated a number of lines, near the end of chapter XVIII, to another legendary object, namely the shield of Achilles, shown in \cite{jung2017}. This richly adorned shield would have been forged by Vulcan at the behest of the hero's mother. While dealing with the shield's manufacture, the poet provides a number of brief narratives, attached to and thus inspired by distinct scenes engraved on the metal surface. Virgil's passage is one of the best known cases of `ekphrasis', a word of Greek origin, that designates a rhetorical artifice, consisting of the written description of a true or legendary work of art \cite{webb2009}. 

Real objects of supreme art are the many sculptures of hands, singly represented, not as part of a body, produced by the French sculptor Auguste Rodin. Some of these hands are meant to suggest God's creative power, and also, in a reduced scale, the artist's productive work. The Austrian poet Rainer Maria Rilke, while serving Rodin as secretary (1905–1906), wrote a book \cite{rilke2006} about the sculptor's masterpieces, not forgetting to mention the all important hands. And, inspired by their image, he composed in German a poem entitled \textit{Herbst} (Autumn) — evoking the season when the leaves of the trees sadly come to fall as a metaphor of our many frustrated dreams, but also when divine hands express a message of hope \cite{rilke1902}:

\blockquote{Und doch ist Einer, welcher dieses Fallen \newline
unendlich sanft in seinen Händen hält. \newline
[But there is One, Who holds what falls \newline
with infinite tenderness in His hands.]}

Another artifice, this time employed by plastic artists in general, is equally named by a word of Greek origin: `Anamorphosis', defined as \enquote{a distorted projection that requires the viewer to occupy a specific vantage point, use special devices, or both to view a recognizable image} \cite{collins1992}. A well-known example is Holbein's painting \textit{The Ambassadors} (1533), which, disguised amid the luxury of the scene, ironically places a human skull, as a remembrance of mortality.\footnote{\href{https://www.nationalgallery.org.uk/paintings/hans-holbein-the-younger-the-ambassadors}{https://www.nationalgallery.org.uk/paintings/hans-holbein-the-younger-the-ambassadors}}  

Drawings and paintings, incidentally, are a practical recourse for psychological analysis, including Rorschach psychological test \cite{exner2002}, Freud's interpretation of a Leonardo da Vinci's painting \cite{freud1910}, and Jung's analyses of his patients' drawings \cite{jacobi1973}. Another practical application that explicitly combines serious information systems with narrative is the fascinating topic of \textit{Data Storytelling} \cite{knaffic2015}.

\section{The ImageTeller Prototype}
\label{sec3}

The proposed prototype, called ImageTeller, is designed to explore the narrative potential of images by employing AI agents, namely GPT-4o, GPT-4o Vision, and Stable Diffusion XL, to analyze visual content and generate stories. Unlike purely text-driven approaches, ImageTeller's integration of visual data aims to produce stories that harness both the imaginative potential of AI and the contextual richness of visual storytelling. The system provides an interactive platform where users can input images from various sources, guiding the AI to generate narratives that are both structurally coherent and creatively compelling. The prototype is accessible through our public website: \href{https://narrativelab.org/imageteller/}{https://narrativelab.org/imageteller/}.

\subsection{User Interface} 
\label{sec31}

\begin{figure}[htb]
	\centering
	\includegraphics[width=1\linewidth]{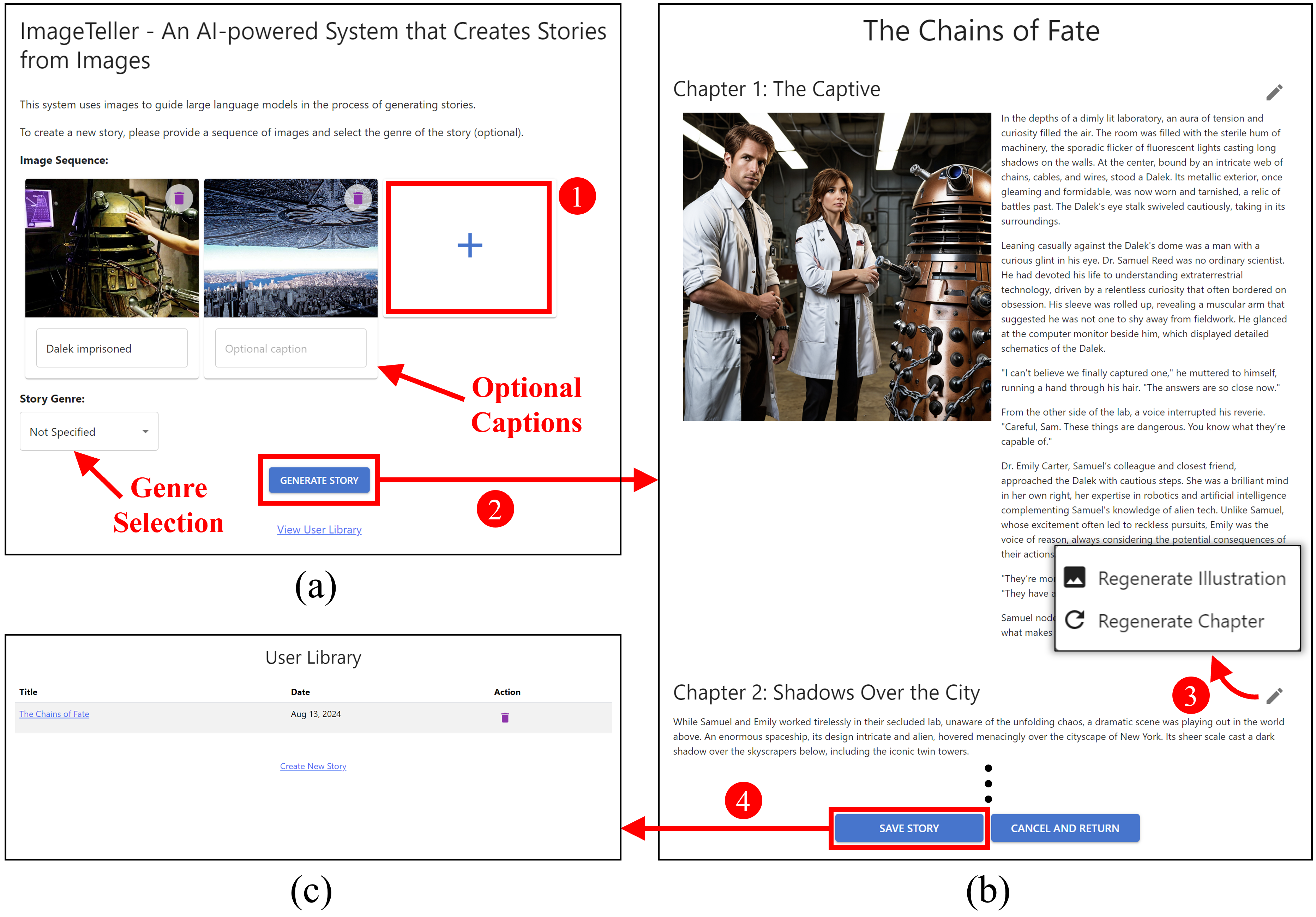}
	\caption{The user interface of ImageTeller.}
	\label{fig2}
\end{figure}

As illustrated in Figure \ref{fig2}, the user interface of ImageTeller is designed to facilitate the process of generating narratives from images. To begin, users can upload images on the system's initial screen (Figure \ref{fig2} (a)) by clicking on the \enquote{+} button (Action 1 in Figure \ref{fig2} (a)). In the example presented in Figure \ref{fig2}, the user has uploaded two images: the first is a scene from the \textit{Doctor Who} series featuring an imprisoned Dalek, and the second is an image from the movie \textit{Independence Day} (1996), showing an alien spaceship over New York. For this example, the caption \enquote{Dalek imprisoned} was entered for the first image, and no specific genre was selected for the story.

Once the images are added and arranged, the user can click on the \enquote{Generate Story} button to initiate the narrative generation process (Action 2 in Figure \ref{fig2} (a)). The system then transitions to the story composition screen (Figure \ref{fig2} (b)), where the user can visualize the generated story (titled \enquote{The Chains of Fate}) and its chapters, each accompanied by a title and an illustration.

For each chapter, the user has the option to either regenerate the illustration or regenerate the entire chapter by using the respective buttons available on the interface (Action 3 in Figure \ref{fig2} (b)). These options allow the user to refine the narrative and its visual representation to better suit their preferences. At the bottom of the screen, the user can click on the \enquote{Save Story} button to save the generated story (Action 4 in Figure \ref{fig2} (b)), which will then be added to the user's personal library (Figure \ref{fig2} (c)). The full story generated for the example presented in Figure \ref{fig2} is available at: \href{https://narrativelab.org/imageteller/\#/story/186}{https://narrativelab.org/imageteller/\#/story/186}.

\subsection{System Architecture} 
\label{sec32}

\begin{figure}[htb]
	\centering
	\includegraphics[width=0.8\linewidth]{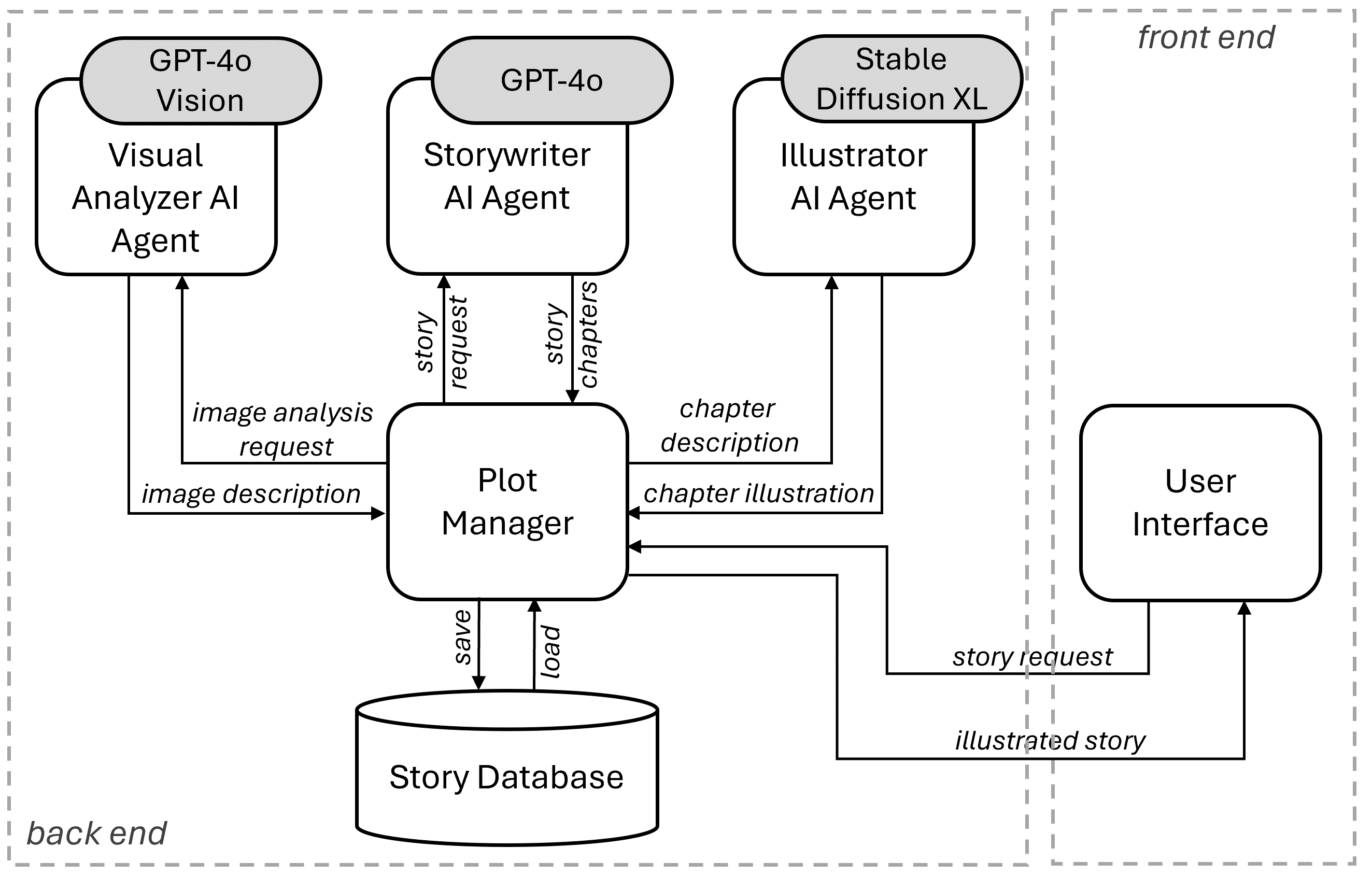}
	\caption{The multi-AI-agent architecture of ImageTeller.}
	\label{fig3}
\end{figure}

As illustrated in Figure \ref{fig3}, the architecture of ImageTeller is based on the multi-AI-agent approach for narrative generation using large language models proposed in our previous works \cite{limaICEC23,limaSBGames23,schetinger2023,lima2024Pattern,limaENTCOM24}, which we adapted in the present work to support the generation of narratives from visual inputs. This architecture incorporates a set of AI agents, each with a distinct role in the narrative generation process: (1) a Visual Analyzer AI Agent, which uses the GPT-4o Vision to interpret and analyze the content of input images; (2) a Storywriter AI Agent, powered by the GPT-4o model, responsible for generating the textual narrative based on the results of the visual analysis; and (3) an Illustrator AI Agent that uses a Stable Diffusion model to create visual representations that complement and enhance the story. The overall process is coordinated by the Plot Manager, which controls the interactions between the AI agents and the narrative structure in a way similar to the plot management strategies used in interactive storytelling systems \cite{delima2023}.

User interaction is facilitated through a user-friendly interface, which allows users to request the generation of stories by providing input images, captions, and selecting a genre for the story. The interface also enables users to visualize the generated narratives in a format that resembles illustrated books, where each chapter of the story is represented by a title, the narrative text, and a visual illustration. Additionally, the system allows for user feedback and iterative refinement of the story, making it possible to request alternative chapters or illustrations, or to restart the narrative generation process.

The AI agents are implemented using a plugin architectural approach, which ensures the flexibility of the architecture for future updates of the AI agents. As indicated in Figure \ref{fig3}, the Visual Analyzer AI Agent explores the image analysis capabilities of GPT-4o Vision, which is accessed via the OpenAI API. The Storywriter AI Agent also utilizes GPT-4o model for text generation, while the Illustrator AI Agent is built upon a fine-tuned version of the Stable Diffusion XL model (Juggernaut XL), hosted on a private server and accessed through a REST API. This modular design not only enhances the system's adaptability but also ensures robust performance and scalability.

\subsection{Image Analysis} 
\label{sec33}

The process of analyzing and describing the visual content of images is a core functionality of ImageTeller. This capability is powered by the advanced vision features of the GPT-4o model, which can extract detailed and contextually rich descriptions from each input image. These descriptions are produced using a prompt designed to extract and generate textual descriptions that are suitable for narrative generation.

The prompt used for image analysis (Prompt \ref{prompt1}) is crafted to guide the model in identifying and describing key elements of the image. Specifically, it instructs the model to focus on the subjects present in the image, their actions, and appearance, as well as any significant interactions and environmental details that contribute to the overall mood and context of the scene. The description is generated as a single, cohesive paragraph to ensure that all relevant aspects are seamlessly integrated into the output.

\begin{prompt}
	\label{prompt1}
	\textit{Analyze the provided image and write a short description of the subjects present in the image, including their actions and appearance. Focus on key elements such as the interactions between subjects and any significant environmental details that contribute to the mood and context of the scene. Write the description as a single paragraph without using headers, bullet points, or numbered items.}
\end{prompt}

When a caption is provided by the user, this additional context is incorporated directly into the analysis. Therefore, the prompt is dynamically adjusted to account for the caption by appending the instruction: \enquote{\textit{When generating the description, take into account that the image has the following caption: <\(C_i\)>}},  where \(C_i\) represents the provided caption for image \(i\). This adjustment allows the model to align the image analysis with the user's desired interpretation, ensuring that the generated description is consistent with any specific narrative cues suggested by the caption. This feature is particularly valuable in scenarios where the user intends to steer the narrative in a particular direction, specify character names, or emphasize certain aspects of the image.

For example, when analyzing Figure \ref{fig1} from this paper, the model produced the description presented in Output \ref{output1}, which captures the scene's key elements and interactions, providing a detailed description that reflects the atmosphere and dynamics of the medieval setting.

\begin{output1}
	\label{output1}
	\textit{The image depicts a medieval scene that appears to be a courtly romance, likely inspired by chivalric literature or noble life. In the center, a man and a woman are engaged in an intimate gesture, with the man gently holding the woman's face while they exchange a kiss. The man is dressed in earthy tones with a gold robe, while the woman is adorned in an elaborate red dress with floral patterns, emphasizing her status and elegance. Her headdress suggests noble or royal rank. On the left, another man dressed in a richly decorated blue robe with gold details looks on, perhaps symbolizing an observer or a companion. The background is a vibrant tapestry of red with intricate geometric patterns, suggesting a luxurious setting. The mood is one of romantic intrigue and noble affection, common themes in medieval art, with the emphasis on delicate gestures and the opulence of their surroundings reinforcing the scene's courtly nature.}
\end{output1}

When the same caption used in this paper is provided (\enquote{First tryst of Guinevere and Lancelot, arranged and observed by Galehaut}), the model produces a more contextually nuanced description (Output \ref{output2}), which not only integrates the specific context provided by the caption but also incorporates additional elements from the model's broader knowledge of the Arthurian legend, enriching the narrative and deepening the analysis.

\begin{output1}
	\label{output2}
	\textit{The image depicts a medieval scene illustrating the first tryst of Guinevere and Lancelot, a significant moment in Arthurian legend. Guinevere, dressed in a richly detailed red dress adorned with floral patterns, is shown intimately close to Lancelot, who is wearing a gold robe. The two are engaged in a tender moment, with Lancelot gently touching Guinevere's face, symbolizing the beginning of their romantic relationship. The atmosphere is intimate, highlighted by their close proximity and the affectionate gestures they share. Observing this interaction is Galehaut, who is dressed in a blue robe decorated with intricate gold details, suggesting his nobility and his role as an orchestrator of this meeting. The background features a patterned red tapestry, adding to the luxurious and courtly setting, which enhances the sense of secrecy and passion in this iconic moment of romance and intrigue. The overall mood of the image conveys the delicate balance between duty and desire, capturing the essence of their forbidden love.}
\end{output1}

\subsection{Story Generation} 
\label{sec34}

The story generation process in ImageTeller is designed to transform the visual and contextual data obtained from the image analysis into a cohesive and engaging narrative. This process is powered by the GPT-4o model, which is guided by a structured prompt system that varies depending on the user's input, particularly regarding the type and genre of the narrative. 

Two types of narratives are handled: story-driven and data-driven. A \textit{story-driven narrative} focuses on traditional storytelling and can either follow the conventions of a specific genre — such as Romance, Tragedy, or Mystery — or be more general, without strict adherence to genre-specific elements. In contrast, a \textit{data-driven narrative} is designed to emphasize the communication of insights and data points. This type of narrative prioritizes clarity and relevance, presenting complex information in an engaging and accessible way. While data storytelling could be considered a genre — and ImageTeller does present it as such to users — it requires a different approach to narrative construction, with specific instructions focused on integrating data insights rather than following traditional story arcs. For this reason, we have separated it from the standard genre options in the story generation process to ensure that the narrative style and structure are appropriately tailored to the unique demands of data-driven storytelling.

In a story-driven narrative, a genre is represented by a pair \(G_i = (g_i^{name}, g_i^{description})\), where \(g_i^{name}\) identifies the name of genre \(G_i\), and \(g_i^{description}\) represents a textual description of genre \(G_i\). In our current prototype, we adopted the conventions of five fundamental genres that we have specified in a previous work \cite{delima2024multigenre}, which are presented in Table \ref{tab1}.

\begin{table}[htb]
	\centering
	\caption{Definitions of the five fundamental genres used by ImageTeller.}
	\renewcommand{\arraystretch}{1.4}
	\label{tab1}
	\begin{tabular}{|l|l|m{13.2cm}|}
		\hline
		\multirow{2}{*}{\(G_1\)} & \(g_1^{name}\) & \textit{Comedy} \\ \cline{2-3} & \(g_1^{description}\) & \textit{The world is just and strict but finally becomes more free and desirable. The protagonist is initially hilariously vain, self-important and aspiring, but at the end conforms to the world's norms.} \\ \hline
		\multirow{2}{*}{\(G_2\)} & \(g_2^{name}\) & \textit{Romance} \\ \cline{2-3} & \(g_2^{description}\) & \textit{The world is just but momentarily disturbed by the occurrence of a villainy. The protagonist performs a heroic adventurous quest.} \\ \hline
		\multirow{2}{*}{\(G_3\)} & \(g_3^{name}\) & \textit{Tragedy} \\ \cline{2-3} & \(g_3^{description}\) & \textit{The world is just but governed by fate and unforgiving. The protagonist misbehaves and finally succumbs and dies.} \\ \hline
		\multirow{2}{*}{\(G_4\)} & \(g_4^{name}\) & \textit{Satire} \\ \cline{2-3} & \(g_4^{description}\) & \textit{The world is not just, it is dystopian, grotesque and absurd. The protagonist is helpless.} \\ \hline
		\multirow{2}{*}{\(G_5\)} & \(g_5^{name}\) & \textit{Mystery} \\ \cline{2-3} & \(g_5^{description}\) & \textit{The world is just but has unknown or unexplained or fantastic elements. The protagonist makes a discovery.} \\ \hline
	\end{tabular}
\end{table}

The prompt system for narrative generation is composed of five modular components that are combined to create the final instruction set for the model. These components are: 

\begin{enumerate}
	\item \textbf{General Narrative Instruction} (\(P_{general}\)), which offers core guidance for generating the story, including the structure, title format, and chapter division. This instruction ensures that the story follows a cohesive narrative flow and effectively brings the visual descriptions to life; 
	\item \textbf{Story-driven Instruction} (\(P_{story}\)), which focuses on transforming the image descriptions into a rich, character-driven narrative. This instruction guides the development of characters, their interactions, and the overall story dynamics, ensuring the narrative captures the essence of the scenes while maintaining a traditional storytelling style;
	\item \textbf{Data-driven Instruction} (\(P_{data}\)), which shifts the focus of the narrative towards communicating insights and connections in a compelling way, tailoring the story to effectively present data-driven content with clarity and engagement; 
	\item\textbf{Genre Specification} (\(P_{genre}\)), which defines the story genre (if selected) and incorporates genre-specific elements such as setting, tone, themes, and character types, ensuring the story adheres to the conventions of the chosen genre; and
	\item \textbf{Image Descriptions} (\(P_{image}\)), which contains the sequentially ordered descriptions of the images generated during the image analysis phase. 
\end{enumerate}

Table \ref{tab2} presents the parameterized text for the five components of the prompt system.

\begin{table}[htb]
	\centering
	\caption{Components of the prompt system for narrative generation, where <\(g_i^{name}\)> identifies the genre name, <\(g_i^{description}\)> represents the definition of the genre, and <\(I_{description}\)> is an ordered set of image descriptions.}
	\renewcommand{\arraystretch}{1.4}
	\label{tab2}
	\begin{tabular}{|l|m{13.9cm}|}
		\hline
		\textbf{Component} & \textbf{Prompt Text} \\ \hline
		\textbf{\(P_{general}\)} & \textit{Using the following set of sequentially ordered image descriptions, write a cohesive, engaging and complete story resembling a book narrative. The story must have a creative title written using markdown header level 1 (\#). Follow the sequence of events as represented in the descriptions, maintaining consistency in mood and context. Ensure the narrative has a clear beginning, middle, and end. Divide the story into a natural number of chapters, each with a name and number, using markdown header level 2 (\#\#) for the chapter titles.} \\ \hline
		\textbf{\(P_{story}\)} & \textit{When writing the story, assign names to the subjects and develop their characters based on their described actions and appearances. Ensure the story incorporates all key elements from the image descriptions, including the subjects, their interactions, and significant environmental details. Focus on creating a narrative and dialogue that brings the story to life, rather than merely describing the images. Exclude any details specific to the image format, such as speech balloons, text boxes, or floating text.} \\ \hline
		\textbf{\(P_{data}\)} & \textit{Highlight key insights, connect the dots between different data points, and present a clear and engaging story. Use metaphors, analogies, and emotional appeal to make the story engaging. Avoid referring directly to the images or describing them. Instead, weave the data into the narrative naturally, as if explaining the insights to someone who has never seen the images. Also suggest potential implications, actions, or reflections based on the insights presented.} \\ \hline
		\textbf{\(P_{genre}\)} & \textit{Incorporate genre-specific elements such as setting, tone, themes, and character types to clearly reflect the <\(g_i^{name}\)> genre. Use the following definition for the <\(g_i^{name}\)> genre: <\(g_i^{description}\)>} \\ \hline
		\textbf{\(P_{image}\)} & \textit{Image descriptions: <\(I_{description}\)>} \\ \hline
	\end{tabular}
\end{table}

The final prompt used to guide the GPT-4o model in the narrative generation process is assembled by combining the components of the prompt system based on the user's preferences.

For a story-driven and genre-specific narrative, the final prompt is defined by:

\[P_{final} = P_{general} + P_{story} + P_{genre} + P_{image}\]

When no genre is specified, the final prompt is composed as:

\[P_{final} = P_{general} + P_{story} + P_{image}\]

For a data-driven narrative, the final prompt is given by:

\[P_{final} = P_{general} + P_{data} + P_{image}\]

Prompt \ref{prompt2} provides an example of the final prompt created for a story-driven narrative in the Tragedy genre, using the image description generated for Figure \ref{fig1} as input (Output \ref{output2}).

\begin{prompt}
	\label{prompt2}
	\textit{Using the following set of sequentially ordered image descriptions, write a cohesive, engaging and complete story resembling a book narrative. The story must have a creative title written using markdown header level 1 (\#). Follow the sequence of events as represented in the descriptions, maintaining consistency in mood and context. Ensure the narrative has a clear beginning, middle, and end. Divide the story into a natural number of chapters, each with a name and number, using markdown header level 2 (\#\#) for the chapter titles. When writing the story, assign names to the subjects and develop their characters based on their described actions and appearances. Ensure the story incorporates all key elements from the image descriptions, including the subjects, their interactions, and significant environmental details. Focus on creating a narrative and dialogue that brings the story to life, rather than merely describing the images. Exclude any details specific to the image format, such as speech balloons, text boxes, or floating text. Incorporate genre-specific elements such as setting, tone, themes, and character types to clearly reflect the Tragedy genre. Use the following definition for the Tragedy genre: The world is just but governed by fate and unforgiving. The protagonist misbehaves and finally succumbs and dies.}

	\textit{Image descriptions:}
	
	\textit{1. The image depicts a medieval scene illustrating the first tryst of Guinevere and Lancelot, a significant moment in Arthurian legend. Guinevere, dressed in a richly detailed red dress adorned with floral patterns, is shown intimately close to Lancelot, who is wearing a gold robe. The two are engaged in a tender moment, with Lancelot gently touching Guinevere's face, symbolizing the beginning of their romantic relationship. The atmosphere is intimate, highlighted by their close proximity and the affectionate gestures they share. Observing this interaction is Galehaut, who is dressed in a blue robe decorated with intricate gold details, suggesting his nobility and his role as an orchestrator of this meeting. The background features a patterned red tapestry, adding to the luxurious and courtly setting, which enhances the sense of secrecy and passion in this iconic moment of romance and intrigue. The overall mood of the image conveys the delicate balance between duty and desire, capturing the essence of their forbidden love.}
\end{prompt}

A key feature of \(P_{general}\) is its directive that instructs the model to generate the narrative in the expected output format, with the story divided into chapters and markdown headers used to identify titles. This structured format ensures the narrative is organized with clear distinctions between the story title, chapter titles, and chapter content. By analyzing the markdown headers, the system can parse the generated narrative and process each chapter individually to produce the final formatted version of the narrative to be displayed to users. Output \ref{output3} shows an example of the raw narrative content generated for Prompt \ref{prompt2}.

\begin{output1}
	\label{output3} 
	
	\
	
	\# \textit{The Forbidden Tryst}
	
	\#\# \textit{Chapter 1: The Meeting}
	
	\textit{In the hallowed halls of Camelot, where the stone walls resonated with the whispers of countless legends, a tale of doomed romance began to unfurl. Lady Guinevere, the epitome of grace and beauty, was married to King Arthur, the mighty and just ruler of the realm. Yet, fate had a different story written in the stars for her heart.}
	
	\textit{Guinevere, adorned in a richly detailed red dress with floral patterns, moved as if each step was a dance. Her beauty captivated all who laid eyes upon her, but it was her spirit, passionate and wild, that truly set her apart. In the heart of the castle, an intimate setting shaped by a grand tapestry of crimson and gold, she found herself ensnared by emotions she scarcely understood.}
	
	\textit{It was here that she met Sir Lancelot du Lac, the king's most trusted knight. Lancelot, cloaked in a robe of shimmering gold, was the paragon of chivalric virtue and unmatched bravery. His loyalty to Arthur was unwavering, yet his heart, treacherous as a tempest, found itself beating to the rhythm of Guinevere's gaze.}
	
	\textit{Watching over this clandestine meeting was Galehaut, a nobleman of towering stature and noble bearing. Draped in a deep blue robe adorned with intricate golden details, he occupied a unique position – both friend and confidant to Lancelot. With eyes that seemed to understand more than they saw, Galehaut brought these two fated souls together, knowing well the peril that their bond would bring.}
	
	\textit{The atmosphere hummed with unspoken words as Lancelot tenderly touched Guinevere's face, a gesture filled with tenderness and longing. It was in this moment, under Galehaut's observant gaze, that their forbidden love was born. Torn between loyalty to her king and the burgeoning passion for Lancelot, Guinevere's heart waged a silent battle.}
	
	\textit{\enquote{Lancelot,} she whispered, her voice trembling with the weight of unspoken emotions, \enquote{what fate is this that binds us so?}}
	
	\textit{\enquote{Milady,} he replied, his voice a cloak of velvet, \enquote{it is a cruel jest that love should find us thus, yet it is a jest I would not trade for all the world.}}
	
	\#\# \textit{Chapter 2: The Unforgiving Fate}
	
	\textit{Days turned into nights, and the seasons changed, but the bond between Guinevere and Lancelot grew ever stronger. Their meetings, always under the watchful eye of Galehaut, became a sanctuary for their hearts. Each stolen glance and secret touch was a defiance of the fate that sought to keep them apart.}
	
	\textit{However, the world governed by fate is unforgiving. Whispers of their romance began to drift through the castle like a cold breeze heralding a storm. The love that once brought them solace now became a harbinger of their undoing. The nobles of Camelot, ever eager for scandal, began to suspect, and soon, these suspicions reached the ears of King Arthur.}
	
	\textit{Arthur, a paragon of justice and a man of unwavering principles, found himself at an unfathomable crossroads. His love for Guinevere was as pure as his trust in Lancelot. The very thought of their betrayal was a blade that cut deeper than any enemy's sword.}
	
	\textit{\enquote{My king,} Merlin, the wise and enigmatic sorcerer of the court, spoke softly, \enquote{the path you must tread is fraught with sorrow. Yet, to ignore this would be to invite chaos.}}
	
	\textit{Arthur's eyes, heavy with the burden of his crown, turned to his trusted advisor. \enquote{Merlin, how does one judge between love and loyalty? How can I mete out justice to those I once called my heart?}}
	
	\textit{Merlin's gaze held a sadness born of centuries. \enquote{The laws of man are pale imitations of the laws of the stars, my king. But remember, it is not the fault of the heart to love, but the hand that acts upon it that must be judged.}}
	
	\#\# \textit{Chapter 3: The Fall}
	
	\textit{As fate would have it, the lovers' secret was no longer theirs to keep. Confronted by Arthur, Guinevere and Lancelot stood amidst the ruins of their once sacred bond. The castle, usually a place of grandeur and festivity, now housed a court of judgment.}
	
	\textit{Arthur's voice, though trembling, bore the weight of betrayal. \enquote{How could you, Guinevere? How could you, Lancelot? My most trusted knight, my beloved queen – you have broken the sanctity of this kingdom.}}
	
	\textit{Guinevere, tears glistening on her cheeks, stepped forth. \enquote{Arthur, my heart did not choose this path lightly. It is a cruel fate that binds us, one that no court or crown may change. But my love for you, though different, remains untainted.}}
	
	\textit{Lancelot, standing resolute despite the turmoil within, spoke with a voice like steel. \enquote{My liege, I have wronged you in ways words cannot amend. Yet, my love for Guinevere is as true as my loyalty once was. I stand ready to face whatever punishment you deem just.}}
	
	\textit{The trial was swift. Galehaut, though heartsick at the sight of his friends' demise, knew there was no swaying the course set by the wheel of fate. Guinevere was sentenced to live out her days in a convent, a stark contrast to the life of royalty she once knew. Lancelot was exiled, stripped of his knighthood and honor.}
	
	\textit{As he was led away, Lancelot turned to Guinevere one last time. \enquote{My love, though our bodies may be separated by oceans and walls, our hearts will forever be one beneath the same sky.}}
	
	\textit{And so, the story of Guinevere and Lancelot, a tale of a love that defied kings and kingdoms, ended in tragedy. Their misdeeds, born not of malice but of a cruel and unforgiving fate, brought them to their downfall. In the annals of Camelot, their story would be whispered as a caution of passion and loyalty, and the dire consequences that follow when they are in conflict.}
	
	\textit{Yet, as the stars continued to twinkle above the ancient castle, it was said that on the quietest of nights, you could hear the whisper of their eternal love, a testament to the enduring power of the heart, even in the face of tragic destiny.}
\end{output1}

\subsection{Image Generation} 
\label{sec35}

In ImageTeller, each story is structured into multiple chapters, with each chapter comprising a title, text, and an illustration. The process for generating illustrations for the chapters takes advantage of the recent advancements in text-to-image machine learning models, such as DALL-E, Midjourney, and Stable Diffusion. In our current implementation, we use the Juggernaut XL model,\footnote{\href{https://civitai.com/models/133005/juggernaut-xl}{https://civitai.com/models/133005/juggernaut-xl}} which is a fine-tuned version of the Stable Diffusion XL model.

The process for generating an illustration starts by using the GPT-4o model to identify and describe a significant event in the text of a chapter. The generated description serves as input for the image generation model. Prompt \ref{prompt3} presents the prompt used to instruct GPT-4o for this task (the parameter \(C_i^{text}\) represents the text of a chapter \(C_i\)).

\begin{prompt}
	\label{prompt3}
	\textit{Identify a single significant event in the provided story chapter and write a short description (maximum of 60 words) to be used as input for a stable diffusion model to generate an illustration for the chapter. Describe the subjects involved in the scene and their appearance, including physical attributes and clothing style and color. Write only the description and do not add any formatting elements in the text. Here is the story chapter: <\(C_i^{text}\)>}
\end{prompt}

While directly providing the generated description to a text-to-image model would suffice for producing an illustrative representation for a chapter, we implemented additional prompt optimizations to improve image quality. First, we incorporated a default illustration style (\enquote{\textit{Hyperdetailed photography with highly detailed textures, accurate lighting, and realistic colors}}), which is combined with the chapter description to define a consistent visual aesthetic for all generated images. Second, we added a general negative prompt to guide the model toward generating more accurate images by explicitly identifying undesirable visual elements in narrative events, as proposed in our previous work \cite{limaENTCOM24}. A third optimization technique implemented in ImageTeller involves the use of parentheses to increase the emphasis on certain elements within the prompt for image generation, ensuring that the model focuses on the most critical aspects of the scene. This approach, as detailed in our previous work \cite{lima2024Pattern}, helps produce more accurate and visually compelling illustrations by guiding the model's attention to the important nouns and adjectives in the chapter description.

Considering chapter 1 of the story presented in Output \ref{output3} as an example, the description of the chapter's significant event generated by GPT-4o is presented in Output \ref{output4}.

\begin{output1}
	\label{output4}
	\textit{In the dimly lit chamber of Camelot, Lady Guinevere, a vision of elegance in a flowing red dress with floral details, stands close to Sir Lancelot, who is draped in a shimmering gold robe. Lancelot gently touches her face, their eyes locked in a tender and forbidden exchange.}
\end{output1} 
	 
After applying the prompt optimization techniques, the final prompt for the text-to-image model is generated as shown in Prompt \ref{prompt4}.

\begin{prompt}
	\label{prompt4}
	
	\
	
	Prompt: \textit{In the ((dimly lit chamber)) of ((Camelot)), Lady ((Guinevere)), a vision of ((elegance)) in a flowing (((red dress))) with (((floral details))), stands close to Sir ((Lancelot)), who is draped in a shimmering (((gold robe))). ((Lancelot)) gently touches her ((face)), their ((eyes)) locked in a tender and forbidden exchange. Hyperdetailed photography with highly detailed textures, accurate lighting, and realistic colors.}
	
	Negative Prompt: \textit{((poorly drawn hands)), ((poorly drawn face)), blurry, ((bad anatomy)), (((bad proportions))), ((extra limbs)), extra fingers, mutated hands, out of frame, (malformed limbs), ((missing arms)), ((missing legs)), ((extra arms)), (((extra legs))), (((long neck)))}
\end{prompt}

\begin{figure}[htb]
	\centering
	\includegraphics[width=0.4\linewidth]{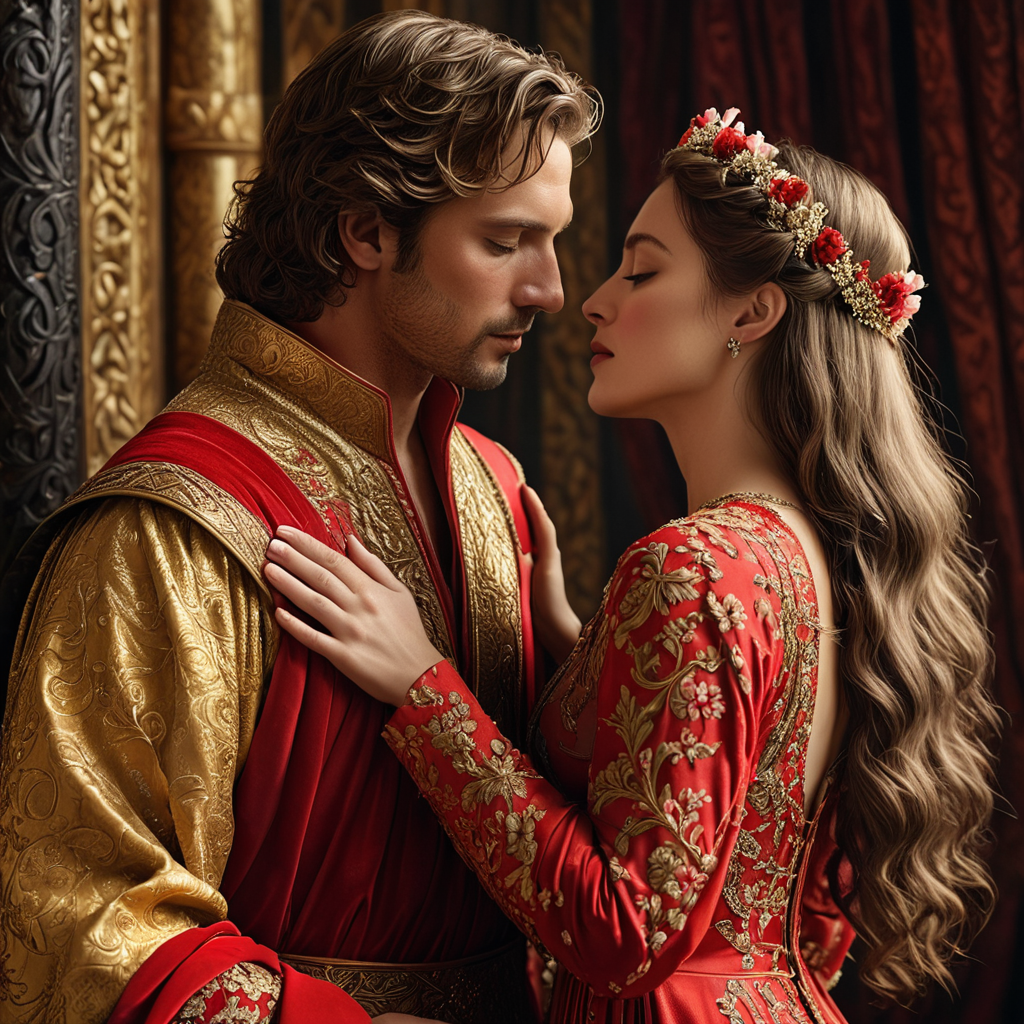}
	\caption{Illustration generated by the Juggernaut XL model for Prompt \ref{prompt4}.}
	\label{fig4}
\end{figure}

Figure \ref{fig4} shows the illustration generated by the Juggernaut XL model using Prompt \ref{prompt4} as input. 

\section{Experiments}
\label{sec4}

\begin{figure}[h]
	\centering
	\includegraphics[width=0.8\linewidth]{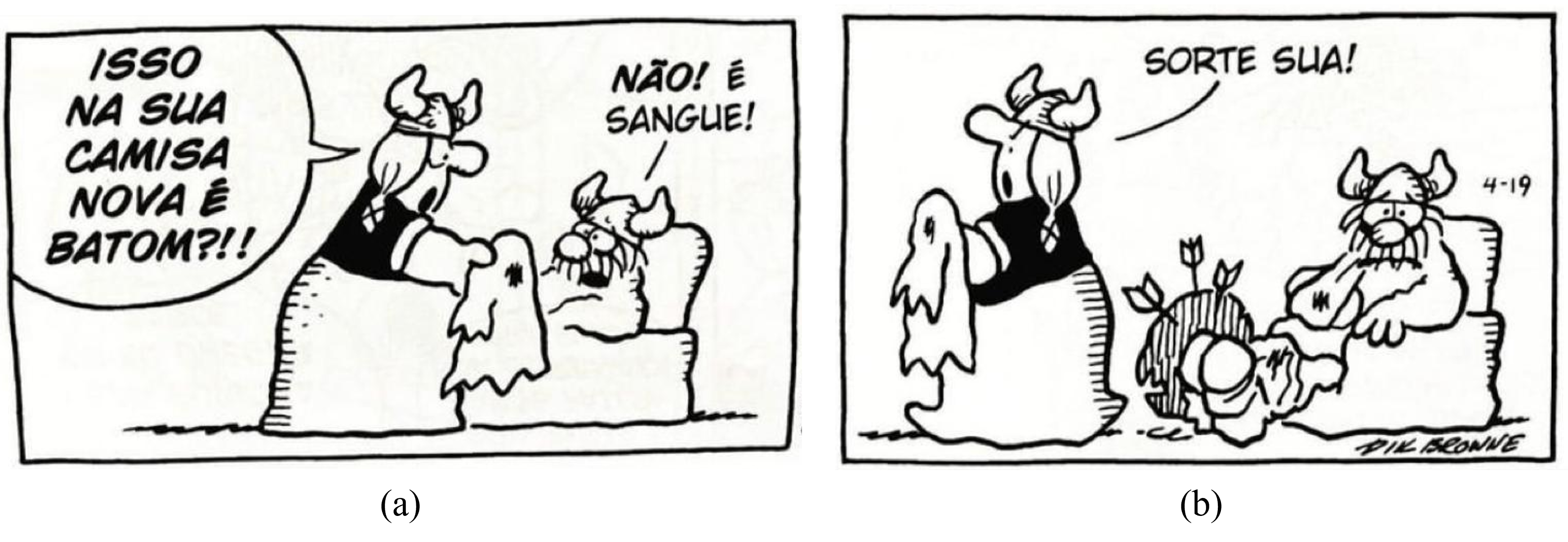}
	\caption{Hägar and Helga \cite{browne19}.}
	\label{fig5}
\end{figure}

To test the capabilities of ImageTeller, we conducted a series of experiments using images of different sources. As a first humorous experiment, we took a two-panel Brazilian version of the \textit{Hägar the Horrible} comic strip, created by Dik Browne and Chris Browne, \footnote{\href{http://www.hagarthehorrible.net/}{http://www.hagarthehorrible.net/}} in which Hägar's wife Helga interrogates the brave Viking about a suspicious red stain on his shirt – \enquote{is it a lipstick stain?}, she asks. When he replies \enquote{no, it is blood}, the irrepressibly stern consort retorts in the second panel \enquote{Lucky you!} (Figure \ref{fig5}). We thought of looking for the original English version but changed our minds as our three trials with the experiment revealed one additional interesting feature of the tool: besides generating the expected funny stories, first with the same Portuguese phrases plus others in the same language as evidence of the AI agent's multilingual ability, secondly with the Portuguese phrases side by side with their English translation, and thirdly solely in English. The complete version of this story can be visualized at: \href{https://narrativelab.org/imageteller/\#/story/197}{https://narrativelab.org/imageteller/\#/story/197}.

\begin{figure}[h]
	\centering
	\includegraphics[width=0.9\linewidth]{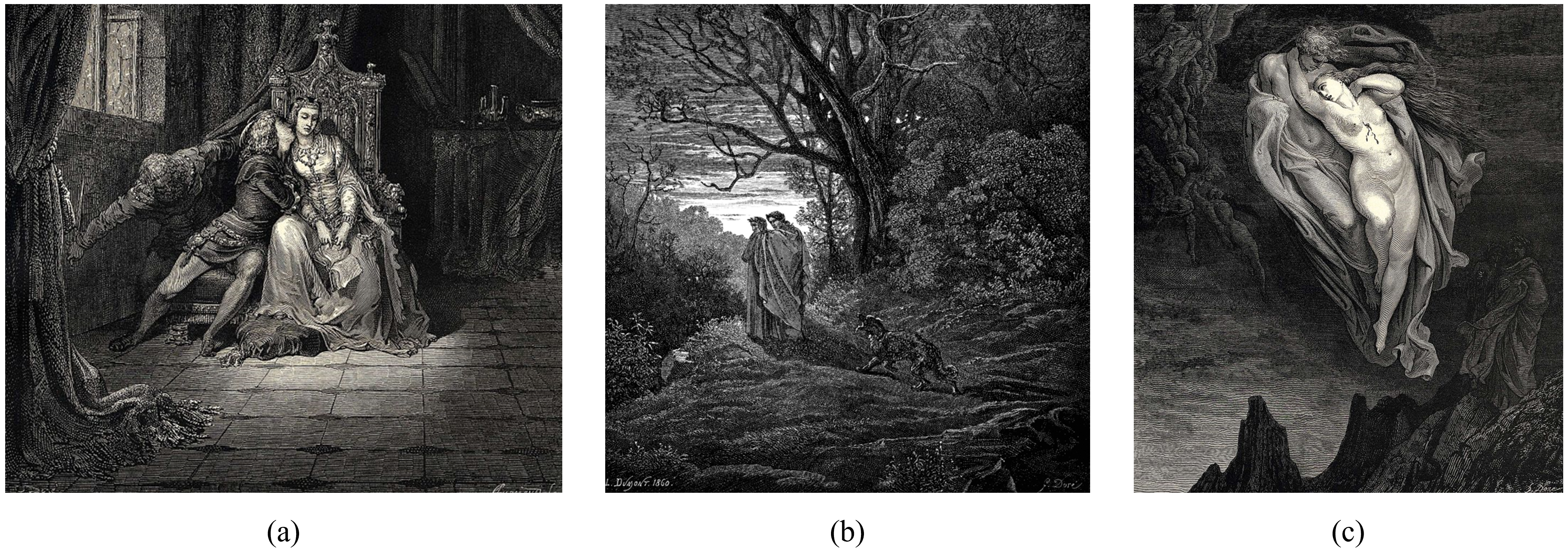}
	\caption{Sequence of image drawn by Gustave Doré \cite{doreinferno61}.}
	\label{fig6}
\end{figure}

We approached another experiment as an occurrence of the tragedy genre \cite{delima2024multigenre}, taken from Dante's \textit{Divine Comedy}. As input, we provided three images drawn by Gustave Doré, whose great art contributed, as we said before, to the moving effect of the Florentine poet's lines. The sequence shows, successively, Francesca da Rimini being kissed by Paulo Malatesta while they were reading together the story of the first tryst of Lancelot and Guinevere (Figure \ref{fig6} (a)), the moment when the Roman poet Virgil came to Dante to serve as guide in the voyage to the supernatural realms (Figure \ref{fig6} (b)), and the two poets watching, in the region of the damned, the air-floating images of the souls of Paolo and Francesca tightly united in a perpetual embrace (Figure \ref{fig6} (c)). A significant feature of Dante's account of the episode is how the soul of Francesca narrates it to the poet, referring to Gahehaut – conceivably inspired by the manuscript's illumination reproduced in our Figure \ref{fig1} – as she blames the book and its author as culprits: \enquote{Galeotto fu il libro e chi lo scrisse [Galehaut was the book and he who wrote it].} \footnote{\href{https://www.goodreads.com/quotes/7505-noi-leggeveamo-un-giorno-per-diletto-di-lancialotto-come-amor}{https://www.goodreads.com/quotes/7505-noi-leggeveamo-un-giorno-per-diletto-di-lancialotto-come-amor}} Appropriate captions were also provided to the system in order to contextualize the images: (a) Francesca da Rimini and Paolo Malatesta reading about the first tryst of Lancelot and Guinevere; (b) Poet Virgil comes to poet Dante in a dark forest to guide him into the realm of the damned souls; and (c) Dante and Virgil contemplate the eternally inseparable souls of Francesca and Paolo. The story generated in this experiment is available at: \href{https://narrativelab.org/imageteller/\#/story/199}{https://narrativelab.org/imageteller/\#/story/199}.

\begin{figure}[h]
	\centering
	\includegraphics[width=1\linewidth]{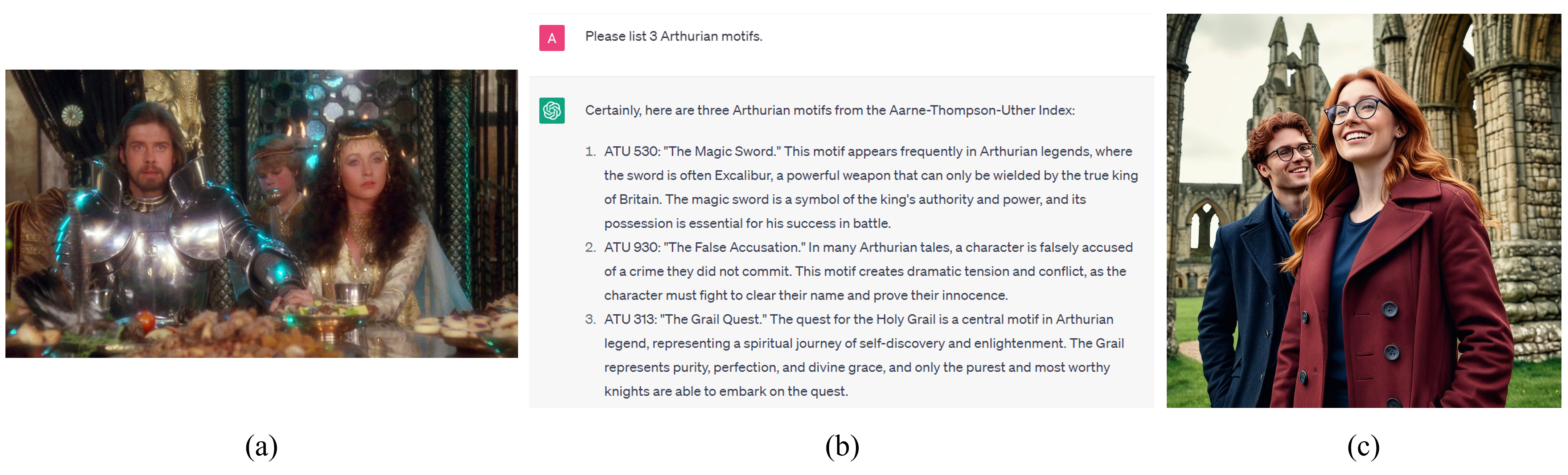}
	\caption{The once and future King and a quest for his legend.}
	\label{fig7}
\end{figure}

In a subsequent experiment, we explored a multiple-source narrative that ImageTeller titled \textit{The False Accusation}, based on images from three widely different origins (Figure \ref{fig7}):

\begin{enumerate}[label=\alph*.]
	\item A scene from the film \textit{Excalibur} (1981), capturing the moment when Queen Guinevere is accused of betraying King Arthur.
	\item A conversation with ChatGPT, where we requested three Arthurian motifs, with \textit{False Accusation} being listed as the second motif.
	\item A scene from a story titled \textit{The Secrets of Glastonbury Abbey}, previously generated by the prototype, in which two students visit the famous abbey.
\end{enumerate}

The third image was retrieved from the User Library of one of the present co-authors by selecting the relevant story and saving the image locally. The complete narrative generated for this experiment is available at: \href{https://narrativelab.org/imageteller/\#/story/213}{https://narrativelab.org/imageteller/\#/story/213}.

An amorous experiment that we tried, curiously suggestive of the `anamorphotic' disguised objects of Section \ref{sec2}, was taken from a scene described in Chrétien's Grail story. A falcon had inflicted a wound on a goose, and three drops of the victim's red blood were visible on the white snow covering the soil. Then Perceval stood marveling at these contrasting colors \cite{chretien2015}:  

\blockquote{Seeing the trampled snow where the goose had lain and the still visible blood, Perceval leaned on his lance to gaze at the image. The blood and the snow together reminded him of the fresh hue on his beloved's face, and he mused until he forgot himself. He thought that the rosy hue stood out against the white of her face like the drops of blood on the white snow. Gazing gave him such pleasure that he believed he was beholding the fresh hue on his beloved's face.}

Since no illustration for this scene was available, we instead provided an image of the textual passage as presented in the book. The resulting narrative is available at:  \href{https://narrativelab.org/imageteller/\#/story/210}{https://narrativelab.org/imageteller/\#/story/210} 

\begin{figure}[htb]
	\centering
	\includegraphics[width=1\linewidth]{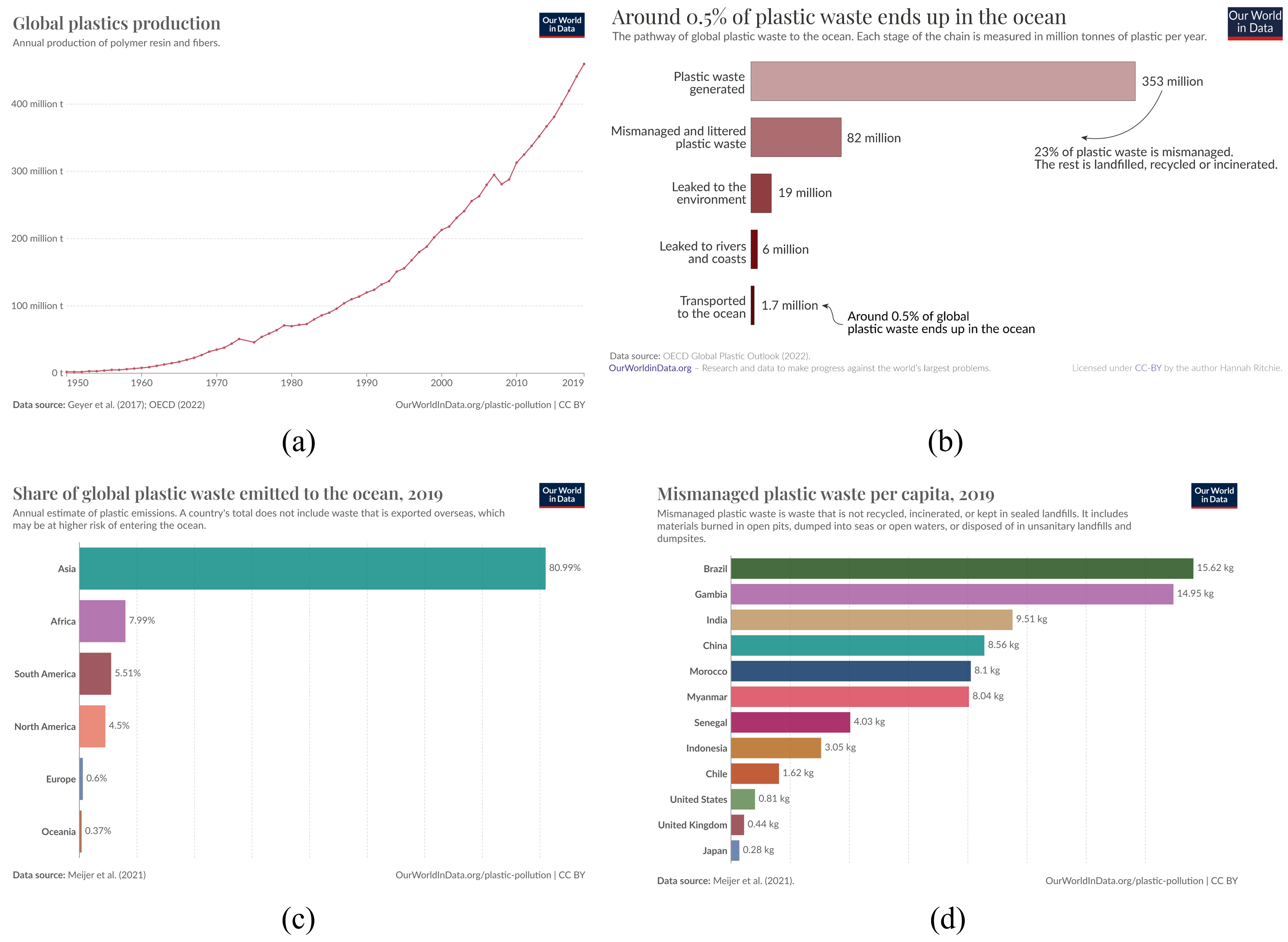}
	\caption{Plastic pollution charts used in the data-driven narrative generation experiment \cite{ritchie23}.}
	\label{fig8}
\end{figure}

To evaluate ImageTeller's capability in generating data-driven narratives, we conducted an experiment using four charts about plastic pollution from the website \textit{Our World in Data} \cite{ritchie23}. These charts provide insights into global plastic production (Figure \ref{fig8}(a)), plastic waste destinations (Figure \ref{fig8}(b)), and the sources of oceanic plastic pollution (Figures \ref{fig8}(c) and (d)). The generated narrative, titled \textit{The Plastic Odyssey: Navigating the Sea of Waste}, balances the factual information from the provided charts with storytelling elements, personifying the environmental impact of plastic pollution while maintaining clarity in presenting key data points. The narrative guides the reader through the historical rise of plastics, the consequences of mismanagement, and the global disparities in pollution contributions. The complete narrative generated for this experiment is available at: \href{https://narrativelab.org/imageteller/\#/story/139}{https://narrativelab.org/imageteller/\#/story/139}.

\section{Concluding Remarks}
\label{sec5}

Our initial experiments, as described in this paper, suggest that analyzing single images or image sequences using the ImageTeller prototype provides a practical and effective approach to generating engaging narratives. The prototype's user-friendly interface, designed with casual users in mind, successfully integrates AI agents to automate the creation of illustrated narrative texts. The simplicity of the interaction, combined with the vast  narrative possibilities, makes the system accessible while still producing creative and coherent stories.

Of particular relevance among the user interaction decisions is the option to prescribe one from a list of fundamental genres, whose traditional conventions would orient the generation process. While it might be argued that letting narratives be guided by old genre conventions would not align with contemporary tastes, our experiments demonstrate that these conventions still hold relevance in narrative composition. This is evident in popular media, such as films and literature, where genre classifications continue to guide audience expectations and engagement. As discussed in our previous work \cite{delima2024multigenre}, genre-based storytelling remains influential in shaping how narratives are constructed and received. Indeed, a recent trend is the return, under a new name, to the original characterization that we have been using of the romance genre:\footnote{\href{https://en.wikipedia.org/wiki/Romantic\_fantasy}{https://en.wikipedia.org/wiki/Romantic\_fantasy}} \enquote{Romantic fantasy or Romantasy is a subgenre of fantasy fiction combining fantasy and romance, describing a fantasy story using many of the elements and conventions of the chivalric romance genre}. At the same time, ImageTeller accommodates users who prefer a more open-ended approach by offering a no-genre option.

Another key feature of the proposed system is its ability to generate data-driven narratives, which expands the system's versatility by enabling the generation of narratives based on visual data representations, such as graphs, charts, and other data visualizations. This capability showcases the potential of AI to translate complex data into coherent and engaging stories that blend information with creative expression. Our experiments with data storytelling illustrate how the system can bridge the gap between traditional narrative forms and modern data communication needs, opening possibilities for broader applications in education, journalism, and business.

As future work, we intend to conduct extensive user experiments aiming at further developing the prototype based on real-world feedback. Such experiments will help us better understand user preferences, improve the generated content's quality, and tailor the system's outputs to meet user needs. Additionally, a promising area for future research is the integration of user modeling techniques to guide the story generation process according to user preferences for narrative content \cite{delima2018a,delima2020} or personality traits \cite{delima2018b}. Finally, we plan to explore ways for improving the control over the generation of illustrations by Stable Diffusion, particularly in ensuring that the traits of the characters remain consistent with the textual descriptions and uniform across successive chapters.

While our current results are promising, continuous iteration and experimentation are essential to fully explore the potential uses of images as input for narrative generation. We encourage readers to explore our system via the links provided in Section \ref{sec3} and share their feedback with us, as all insights are valuable for our ongoing research.

\bibliographystyle{abbrvnat}
\bibliography{references}

\end{document}